%% file: neurips_2025.tex
\documentclass{article}

\usepackage[preprint]{neurips_2025}

\usepackage[utf8]{inputenc} 
\usepackage[T1]{fontenc}    
\usepackage{hyperref}       
\usepackage{url}            
\usepackage{booktabs}       
\usepackage{amsfonts}       
\usepackage{nicefrac}       
\usepackage{microtype}      
\usepackage{xcolor}         
\usepackage{graphicx}
\usepackage{colortbl}
\usepackage{array}
\usepackage{algorithm}
\usepackage{algpseudocode}  
\usepackage{amsmath}
\usepackage{amssymb}
\usepackage{subfig}
\usepackage{listings}
\usepackage{framed}

\usepackage{natbib}
\newcommand\circled[1]{\raisebox{.9pt}{\textcircled{\raisebox{-.9pt}{#1}}}}

\newcommand\nop[1]{}

\input{packages}

\input{macros}
\newcommand{\methodname}{Agent-RLVR\xspace}
\newcommand{\methodnamesft}{Agent-SFT\xspace}

\newcommand{\passatone}{\textsc{Pass}@1\xspace}
\newcommand{\passatk}{\textsc{Pass}@k\xspace}

\title{\methodname: Training Software Engineering Agents via Guidance and Environment Rewards}

\author{
  Jeff Da \qquad Clinton Wang \qquad Xiang Deng \AND Yuntao Ma \qquad Nikhil Barhate \qquad Sean Hendryx \\
  \\
  Scale AI
}

\begin{document}

\maketitle

\input{sections/0-abstract-xiang}
\input{sections/1-introduction}

\input{sections/2-method}
\input{sections/3-experiments}
\input{sections/4-impacts}
\input{sections/5-relatedwork}
\input{sections/6-conclusion}

\clearpage
\bibliographystyle{unsrt}
\bibliography{biblio}


\clearpage


\newpage
\appendix
\input{sections/appendix}

\end{document}

%% file: packages.tex
\usepackage{wrapfig}
\usepackage{amsthm}
\usepackage{placeins}

\usepackage{cite}
\usepackage{amsmath,amssymb,amsfonts}
\usepackage{bm}
\usepackage{pifont}
\usepackage{mathrsfs}
\usepackage{microtype}

\usepackage{graphicx}
\usepackage{textcomp}

\usepackage{mathtools}
\usepackage{multirow}
\usepackage{diagbox}
\usepackage{makecell}
\usepackage{setspace}

\usepackage{booktabs}
\usepackage{colortbl}

\usepackage{url}            
\usepackage{nicefrac}       
\usepackage{xcolor}         
\usepackage{bbm}

\usepackage{dsfont} 

\usepackage{float,wrapfig}

\usepackage{array}
\usepackage{adjustbox}
\usepackage{multirow}
\usepackage{colortbl}
\usepackage{caption}
\usepackage{amsmath}
\usepackage{multicol}
\usepackage{mathtools}

\usepackage{soul}
\usepackage{enumitem}
\usepackage{xspace}

\usepackage{hyperref}       
\usepackage{subcaption}

\usepackage{listings}
\usepackage[capitalize,noabbrev]{cleveref}

\crefname{algorithm}{Alg.}{Algs.}
\Crefname{algorithm}{Algorithm}{Algorithms}
\crefname{appendix}{App.}{App.}
\Crefname{appendix}{Appendix}{Appendices}
\crefname{figure}{Fig.}{Figs.}
\Crefname{figure}{Figure}{Figures}
\Crefname{section}{Section}{Sections}
\crefname{section}{Sect.}{Sect.}
\crefname{subsection}{Sect.}{Sect.}
\Crefname{subsection}{Section}{Sections}
\crefname{subsubsection}{Sect.}{Sect.}
\Crefname{subsubsection}{Section}{Sections}
\crefname{table}{Table}{Tables}
\Crefname{table}{Table}{Tables}

%% file: sections/0-abstract-xiang.tex
\begin{abstract}
   Reinforcement Learning from Verifiable Rewards (RLVR) has been widely adopted as the de facto method for enhancing the reasoning capabilities of large language models (LLMs) and has demonstrated notable success in verifiable domains like math and competitive programming tasks. 
   However, the efficacy of RLVR diminishes significantly when applied to agentic environments. These settings, characterized by multi-step, complex problem solving, lead to high failure rates even for frontier LLMs, as the reward landscape is too sparse for effective model training via conventional RLVR.
   In this work, we introduce \methodname, a framework that makes RLVR effective in challenging agentic settings, with an initial focus on software engineering tasks.
   Inspired by human pedagogy, \methodname~introduces \textit{agent guidance}, a mechanism that actively steers the agent towards successful trajectories by leveraging diverse informational cues. These cues, ranging from high-level strategic plans to dynamic feedback on the agent's errors and environmental interactions, emulate a teacher's guidance, enabling the agent to navigate difficult solution spaces and promotes active self-improvement via additional environment exploration. In the \methodname~training loop, agents first attempt to solve tasks to produce initial trajectories, which are then validated by unit tests and supplemented with agent guidance. Agents then reattempt with guidance, and the agent policy is updated with RLVR based on the rewards of these guided trajectories. We curated a dataset of 817 training environments with problem statements, environments, and guidance in the software engineering domain.
   \methodname~elevates the \passatone~performance of Qwen-2.5-72B-Instruct from 9.4\% to 22.4\% on \textsc{SWE-Bench Verified}.
  We find that our guidance-augmented RLVR data is additionally useful for test-time reward model training, shown by further boosting \passatone~to 27.8\%.
  \methodname~lays the groundwork for training agents with RLVR in complex, real-world environments where conventional RL methods struggle.
\end{abstract}

%% file: sections/1-introduction.tex
\input{figures/figure1}

\section{Introduction}
Reinforcement Learning from Verifiable Rewards (RLVR) has emerged as a prevalent method to train language models for reasoning tasks. Recent models such as OpenAI's o1 and DeepSeek-R1 \citep{DeepSeekAI2025DeepSeekR1IR} have used RLVR to achieve state-of-the-art performance on math and coding tasks. Relying only on reward signals from problem graders or unit test feedback, RLVR yields exceptional improvements in performance along with generalization to related tasks.

However, equivalent success has not been achieved by RLVR in agentic settings. These settings require reasoning across multiple turns, navigating complex problem statements with sequential decision-making, and interacting with external environments via tool use or other interfaces. In these contexts, the probability of generating a correct trajectory across multiple attempts diminishes significantly. As well, interacting with the environment requires additional infrastructure setup and execution time, adding further complexity to training. These complications make it difficult for RLVR to perform well in agentic settings.

To address these issues, we introduce \methodname, a framework for training language model agents that relies solely on rewards produced by the agent-environment interaction. We demonstrate the success of the method in software engineering (SWE) tasks. We choose this setting for several reasons: the setting has objectively verifiable solutions (e.g. passing unit tests), previous work such as SWE-Bench \citep{chowdhury2024swebenchverified} has provided standardized evaluation infrastructure, and advances in this area offer significant practical value for improving developer productivity. The core of our method is the incorporation of \emph{agent guidance}---a multi-agent framework where teacher guidance, which provides targeted information about failing code patches or test expectations, is incorporated during RL training. These hints steer the agent towards successful trajectory by providing these diverse information clues, similar to how a junior software engineer may receive guidance from a teacher or tech lead when first exploring a new codebase.

\nop{To summarize, the key contributions of our work are as follows: \textbf{(1) We propose \methodname, a framework to enable agentic RLVR training.} \methodname addresses the fundamental challenge of sparse rewards in multi-step reasoning environments by incorporating pedagogical elements that guide agents through complex solution spaces without compromising the verifiability of the ultimate reward signal. \textbf{(2) We curate a dataset of SWE tasks that includes problem statements, environments, and expert guidance designed for \methodname.} This dataset goes beyond traditional input-output pairs by capturing complete coding environments with integrated guidance signals, providing a rich resource for training SWE agents. \textbf{(3) We empirically demonstrate that \methodname significantly improves SWE agent performance on SWE-bench Verified.} Our method increases \passatone from 9.4\% to 22.4\% with just 812 training examples. This dramatic improvement with a relatively small training dataset validates the efficiency of our approach and its ability to help agents learn complex multi-step reasoning processes. \textbf{(4) We show that our approach enables effective test-time scaling}. Our RLVR training data applied as a reward model additionally improves \textsc{Pass}@32 performance from 34.2\% to 38.4\%, and supports training a reward model that further enhances \passatone to 27.8\%. These results demonstrate the synergistic benefits of combining our guided RLVR approach with other advanced techniques, establishing a foundation for future research in training agents for complex real-world environments.}
To summarize, the key contributions of our work are as follows:
\textbf{(1) We propose \methodname, a framework to enable effective agentic RLVR training.} \methodname addresses the fundamental challenge of sparse rewards in multi-step reasoning environments by incorporating pedagogical elements that guide agents through complex solution spaces without compromising the verifiability of the ultimate reward signal.
\textbf{(2) We curate a dataset of SWE tasks that includes problem statements, environments, and expert guidance designed for \methodname.} This dataset goes beyond traditional input-output pairs by capturing complete coding environments with integrated guidance signals, providing a rich resource for training SWE agents.
\textbf{(3) We empirically demonstrate the improvement of \methodname~on SWE agent performance.} On SWE-bench Verified, our method increases \passatone from 9.4\% to 22.4\% with just 817 training environments. This dramatic improvement with a relatively small training dataset validates the efficiency of our approach and its ability to help agents learn complex multi-step reasoning processes. We show that guidance is a critical component, as the guidance model improves over the guidance in both \textsc{Pass}@1 (19.8\% $\rightarrow$ 22.4\%) and \textsc{Pass}@32 (34.2\% $\rightarrow$ 38.4\%). Finally, we show that the RLVR data has additional utility for reward model training, as using a reward model trained on the same RLVR data to rank $k=32$ patch generations increases the \passatone~to 27.8\%. These results demonstrate the synergistic benefits of combining our guided RLVR approach with other advanced techniques, establishing a foundation for future research in training agents for complex real-world environments.

%% file: figures/figure1.tex
\begin{figure}[ht]
  \centering
  \includegraphics[width=\textwidth]{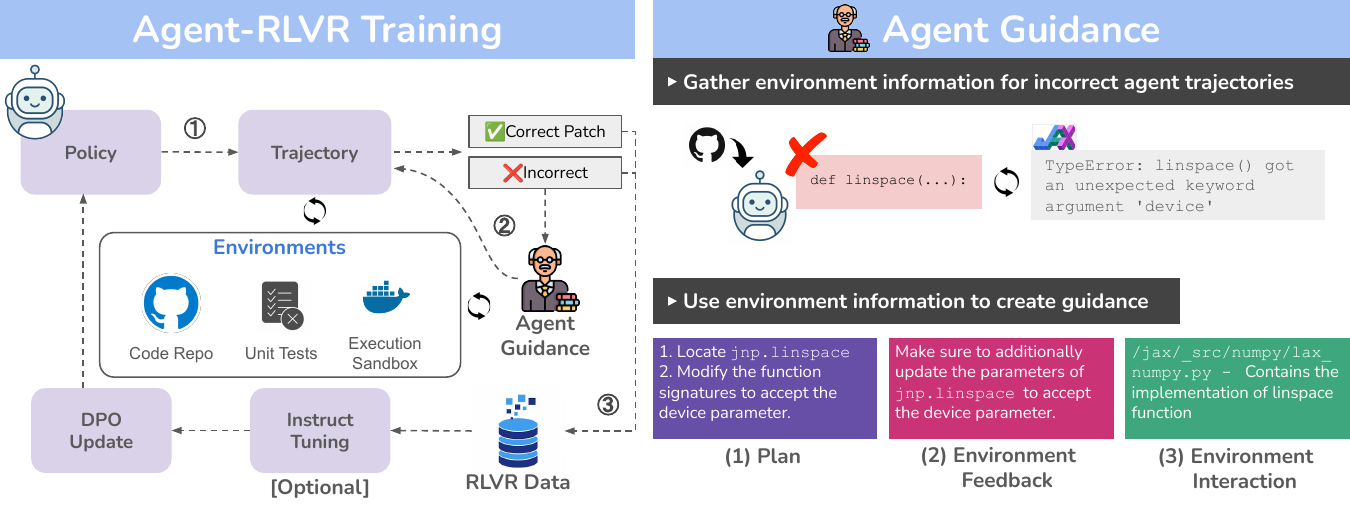}
  \caption{\methodname is a framework for training agents with RLVR using environment feedback and guidance. \circled{1} The agent attempts the problem without any additional guidance and the environment runs unit tests on the generated patch to determine the correctness. \circled{2} We generate guidance for failed patches by leveraging environment information. We provide several types of guidance -- a plan, environment feedback, and environment interaction. The agent reattempts the problem with guidance. \circled{3} Positive trajectories are sampled for instruct-tuning, and trajectories are then used for RLVR to update the agent policy via offline DPO in an iterative manner.}
  \label{fig:figure1}
\end{figure}

%% file: sections/2-method.tex
\section{\methodname: An RL Framework for SWE Agents via Guidance and Environment Rewards}

We introduce \methodname, a framework for training agents via RLVR. \methodname~encourages agents to explore environments more effectively via agent guidance, a method of steering agents towards correct trajectories on challenging tasks. Figure \ref{fig:figure1} shows an illustration of the \methodname~training loop. Note that experiments in this paper describe one iteration cycle.

\subsection{Background: Reinforcement Learning from Verifiable Rewards}
\nop{RLVR is typically preceded by supervised fine-tuning, where a pre-trained large language model (LLM) is fine-tuned on high-quality data for various tasks of interest such as question answering and text summarization, similar to the instruct-tuning done in Section \ref{sec:experiments}}

In RLVR, prompts $x$ are fed to an LLM $\pi_\theta$ that produces responses $y$. A reward function $r(x,y)$ judges prompt-response pairs and provides feedback to the LLM. Following prior works \citep{Stiennon2020LearningTS, Ouyang2022TrainingLM}, the reward function is modulated by a KL divergence penalty that prevents the policy from drifting too far from the reference policy $\pi_{\text{ref}}$ (typically the LLM before RLVR began):

\begin{equation}
R(x,y) = r(x,y) - \beta \log\left[\frac{\pi_{\theta}(y|x)}{\pi_{\text{ref}}(y|x)}\right]
\end{equation}

where $\beta$ is a hyperparameter. LLM parameters $\theta$ are optimized to maximize this expected reward using an algorithm like proximal policy optimization (PPO) or direct policy optimization (DPO) \citep{Rafailov2023DirectPO}. DPO removes the need to explicitly model the reward function, and instead samples pairs of winning responses $y_w$ (that produce higher rewards) and losing responses $y_l$ (that produce lower rewards) for the same input $x$. The DPO loss function drives the LLM towards preferred responses:






\begin{equation}
\mathcal{L}_{\text{DPO}}(\theta) = -\mathbb{E}_{(x,y_w,y_l)\sim\mathcal{D}}\left[ \log \sigma \left(\beta\left(\log\frac{\pi_\theta(y_w|x)}{\pi_{\text{ref}}(y_w|x)} - \log\frac{\pi_\theta(y_l|x)}{\pi_{\text{ref}}(y_l|x)} \right) \right) \right]
\end{equation}

where $\sigma$ is the sigmoid function. In \methodname, we leverage the environment as a source of verifiable feedback for RLVR. In this approach, the preference data $\mathcal{D}$ consists of code solution pairs $(y_w, y_l)$ where $y_w$ passes all unit tests and $y_l$ fails one or more unit tests. 


\subsection{Extending RLVR to Agentic Tasks with \methodname}

While RLVR has shown to be successful at math and competitive coding tasks \citep{DeepSeekAI2025DeepSeekR1IR}, there are several challenges in adapting RLVR to agentic settings. \textbf{Environment complexity.} For math problems, it's possible to grade the answer via a simple string grader or fine-tuned LM grader. However, in agentic settings, an environment takes a significant infrastructure load and execution time to give a reward signal. \textbf{Multi-turn.} Agentic problems are multi-turn, requiring the model to interact with the environment in a series of steps. \textbf{Problem difficulty.} Agentic tasks can be very complex, requiring the model to complete a series of steps in which the agent might never succeed. While these problems are evergreen for agentic tasks, \methodname~helps to mitigate the issues: generated guidance assists the agents in difficult problems and helps steer the model throughout multiple turns, and (by showing the effectiveness of offline RL) our framework enables provided environments to be run asynchronously to speed up reward generation.

In this section, we describe the training dataset, problem formulation, guidance generated, and training loop needed to execute \methodname. In Section \ref{sec:problem_formulation} and Section \ref{sec:guidance}, we describe the format of the problem statements and guidance needed for \methodname. In Section \ref{sec:dataset} we describe our training dataset, and in Section \ref{sec:training} we describe the \methodname~training loop.


\input{figures/algo1}

\subsubsection{Problem Formulation}
\label{sec:problem_formulation}
\textbf{Issue description.} Each task presents the agent with a detailed GitHub issue description that outlines a bug report or feature request from real-world open-source projects. These descriptions often include error messages, expected behavior, actual behavior, and sometimes steps to reproduce the issue. The agent must carefully analyze this natural language problem statement to understand the underlying technical issue, identify relevant parts of the codebase, and formulate a strategy for implementing a solution.

\textbf{Repository.} The agent is given access to a complete snapshot of a real-world Python codebase drawn from popular open-source projects. The agent must efficiently navigate the codebases and locate relevant code components. We are careful not to collect any issues from any repository in the SWE-Bench test set, to avoid any potential data contamination issues.

\textbf{Environment.} The agent operates within a fully functional Docker container that provides an interactive development environment mirroring real-world software engineering workflows. The environment allows the agent to execute code, run specific modules, and observe runtime behavior to better understand the issue. This is similar to other works where unit tests are executed e.g. for coding challenges \citep{Austin2021ProgramSW, Hendrycks2021MeasuringCC}, however in our case, the infrastructure setup incurs additional complexity as each problem has it's own runtime environment. 

\textbf{Unit tests.} Each task includes test suites that serve as the ultimate validation mechanism for the agent's solution. These tests include both regression tests that verify existing functionality remains intact, as well as focused tests that specifically check whether the issue has been resolved. These tests are used to provide verifiable feedback to the agent in RL training.

\subsubsection{Guidance}
\label{sec:guidance}
At the core of \methodname is the introduction of guidance -- a method of assisting the agent in the trajectory rollout stage. During guidance generation, the annotator receives access to the problem statement, patch, and (in the case of environment feedback guidance) the previous trajectory and environment feedback produced by the agent. Although the model is not required to use the guidance, it can help the model to generate the correct trajectory, especially for more difficult problem statements. This promotes active self-improvement for the model, as it is encouraged to explore additional trajectories that it otherwise would not have generated. For this work, we use an external LLM in place of human-generated guidance due to cost considerations. Note that guidance is included during train-time only, at test-time, the inference is the same with and without guidance-trained models. We show that guidance reduces reward landscape sparsity and improves agent \passatk~in Table \ref{tab:guidance_results} and Figure \ref{fig:graph_alt}.
We include the following types of guidance.

\textbf{Plan.} We include guidance to assist the agent in solving the given problem. Derived from the problem statement, the plan is either a series of suggested steps to solve the problem and/or a pointer regarding the crux of the problem. Similar to other works finding that the plan helps to improve generation diversity \citep{Wang2024PlanningIN}, we hypothesize that the plan encourages the agent to explore additional trajectories beyond the range of it's initial policy.

\textbf{Environment feedback.} Given the feedback from the environment, the annotator (or external model) creates guidance that corrects the previous mistake. This aims to guide the agent to reconsider a previous trajectory in which the patch caused an error or specific test to fail.

\textbf{Environment interaction.} We include environment interaction following Learn-by-Interact \citep{Jimenez2023SWEbenchCL}. In particular, we guide the annotator or model to produce guidance with respect to the correct file and patch location. This is to help mitigate potential issues with agent navigation during the localization phase.

\subsubsection{Training Dataset}
\label{sec:dataset}
We create a training dataset for \methodname~complete with environments and guidance generations. The dataset is a set of 817 agent environments for SWE tasks. Given a codebase along with an issue to be solved, the agent is tasked with editing the codebase to address this issue. This requires the agent to navigate the given codebase, coordinate changes across multiple functions, classes, and files, and complete complex reasoning while interacting with execution environments. We include issues from SWE-Gym \citep{Pan2024TrainingSE}, and source additional ones using a similar scraping and annotation pipeline. In total, we include 27 different repos, with 593 problems from SWE-Gym and 219 self-collected problems. The problems are then fed into the \methodname pipeline to collect trajectories. The final training dataset consists of 8186 trajectories in total, out of which 4093 are positive trajectories. Instruct-tuning is done on a subset (20\%) of positive trajectories and the same set of problems is used for instruct-tuning and DPO.

\subsubsection{\methodname~Training Loop}
\label{sec:training}

We describe the training loop for \methodname. First, agent trajectories are generated using a scaffold. In our experiments, we use the scaffold described in Section \ref{sec:experiments}. If the trajectory successfully generates a patch, it is validated by executing the Docker image and running the unit tests. For failing trajectories, where either the unit tests fail or the patch is not generated, the problem is reattempted with guidance. At this stage, the guidance is generated and the agent reattempts the problem with guidance and the new set of generated patches is graded. We find that the guidance augmentation enables the agent to complete more trajectories, as shown in Table \ref{tab:guidance_results}. For each correct trajectory, an incorrect (non-guidance) trajectory for that problem is sampled to create a pair of trajectories for DPO training. If the model has not yet been fine-tuned, the model goes through instruct-tuning where a subset of 20\% of the (non-guidance augmented) trajectories are sampled for SFT training. Finally, DPO training occurs using each trajectory pair as training data, concluding the training iteration. The algorithm is illustrated in Algorithm \ref{algo}.

%% file: figures/algo1.tex
```latex
\begin{algorithm}
\caption{\methodname}
\begin{algorithmic}[1]
\State \textbf{Input:} $M_\theta$ = agent model, $T$ = teacher, $\mathcal{D}$ = dataset of issue descriptions, repositories, and test suites,
$\mathcal{T}_{correct}, \mathcal{T}_{incorrect} = \emptyset$: correct and incorrect agent trajectories
\State Initialize $\mathcal{D}_{RLVR} = \emptyset$
\State \textbf{\circled{1} Initial agent attempt}
\For{$d_i$ in dataset $\mathcal{D}$}
   \State Generate agent trajectory $t_i$: $t_i = M_\theta(d_i)$
   \State Generate reward $r_i$: $r_i = \text{evaluate}(d_i, t_i)$
   \If{$r_i = 1$}
       \State $(d_i, t_i)$ $\rightarrow$ $\mathcal{T}_{correct}$
   \Else
       \State $(d_i, t_i)$ $\rightarrow$ $\mathcal{T}_{incorrect}$
   \EndIf
\EndFor
\State Add to $\mathcal{D}_{RLVR}$: $\{(d_i, t_{i,j}^+, t_{i,k}^-) : r_{i,j} > r_{i,k}\}$ \Comment{Generate preference pairs from multiple rollouts}
\State \textbf{\circled{2} Generate agent guidance}
\For{$(d_i, t_i)$ in $\mathcal{T}_{incorrect}$}
   \State Generate teacher guidance $g_i$: $g_i = T(d_i, t_i)$
   \State Generate new solution with guidance: $t_i' = M_\theta(d_i, g_i)$
   \State Evaluate new solution: $r_i' = \text{evaluate}(d_i, t_i')$
   \If{$r_i' = 1$} 
       \State Add $(d_i, t_i', t_i)$ to $\mathcal{D}_{RLVR}$
   \EndIf
\EndFor
\State \textbf{\circled{3} Update policy}
\State $\mathcal{D}_{SFT} \subset \mathcal{D_{RLVR}}$
\State $\mathcal{L}_{\text{SFT}}(\theta) = -\frac{1}{|\mathcal{D}_{\text{SFT}}|}\sum_{(x_i, r_i, y_i^*) \in \mathcal{D}_{\text{SFT}}} \log p_\theta(y_i^* | x_i, r_i)$
\State $\theta_{\text{SFT}} = \arg\min_\theta \mathcal{L}_{\text{SFT}}(\theta)$
\State $\pi_{\text{ref}} = M_{\theta_{\text{SFT}}}$
\State $\mathcal{L}_{\text{DPO}}(\theta) = -\mathbb{E}_{(x,y_w,y_l)\sim\mathcal{D}_{RLVR}}\left[ \log \sigma \left(\beta\left(\log\frac{\pi_\theta(y_w|x)}{\pi_{\text{ref}}(y_w|x)} - \log\frac{\pi_\theta(y_l|x)}{\pi_{\text{ref}}(y_l|x)} \right) \right) \right]$
\State $\theta_{\text{RLVR}} = \arg\min_\theta \mathcal{L}_{\text{DPO}}(\theta)$
\State \Return $M_{\theta_{\text{RLVR}}}$
\end{algorithmic}
\label{algo}
\end{algorithm}
```

%% file: sections/3-experiments.tex
\section{Experiments}

\subsection{Setup} \label{sec:experiments}

\textbf{Dataset.} We use the aforementioned training dataset for RLVR training, with a total of 817 environments. For guidance generation, we use an external LLM (claude-3-7-sonnet-20250219). We include both on-policy (trajectories generated by Qwen2.5-72B-Instruct) and off-policy (trajectories generated by the LLM used for guidance) data.

\textbf{Trajectory collection.} We sample from the model multiple times to collect trajectories, similar to works in which \passatk~is distilled into \passatone~\citep{Zelikman2022STaRBR, Gulcehre2023ReinforcedS}. Our dataset of 8186 trajectories is generated from sampling from 16 trajectories for each problem during the guidance and non-guidance trajectory sampling. 20\% of the dataset is used for instruct-tuning, where we adapt the model to the task, and the entire dataset is use for DPO training.

\textbf{Model training.} For model training, we use the Qwen-2.5 family of models. We reserve a portion of the dataset for an instruct-tuning phase, such that the model is adapted to learn the format of the task. All models are instruct-tuned for five epochs with a learning rate of 1e-5, and sequence length of 8k. For RLVR training, we use DPO with a learning rate of 1e-6 for 1 epoch.

\textbf{Evaluation dataset.} We evaluate our models on the verified portion of SWE-Bench \citep{chowdhury2024swebenchverified}. We use the evaluation harness provided by the authors\footnote{\url{https://github.com/SWE-bench/SWE-bench}}, and use Modal as our infrastructure stack during evaluation. Several environments in SWE-Bench Verified have been noted to be broken by other works. We found similar issues in our experiments, where 6 instances could not be compiled. Nevertheless, we include all instances in our metrics to remain consistent with other methods.

\textbf{Scaffolding.} We use a basic scaffolding based off Agentless \citep{Xia2024AgentlessDL}, which is widely used for SWE agent evaluation. We pick Agentless for its popularity and maturity at the time of publication. To simplify the scaffold for our experiments, we make some minor modifications and only kept the localization and repair steps. We removed the patch selection step that require rather expensive test generation and execution, so our scaffold is significantly simpler, faster, and less compute intensive than Agentless.
\nop{In our case, the scaffold prompts the agent to perform localization and repair steps. First, the model performs a localization step in which the model is given a skeleton of the repository and is tasked with selecting a list of files in which the relevant edits would take place. For each selected file, the model is also given a list of methods to edit. In the repair step, the model proposes a patch to solve the pull request.}
During the self-improvement stage with guidance, we include an extra prompt in which the model is given access to the additional hints for localization and repair.

\textbf{Metrics.} We report \passatone to compare similar methodologies and for ablation studies, as well as \textsc{Best}@1 to compare with frontier models and other open source models on SWE-Bench. For \passatone, we use greedy decoding at all steps. For \passatk and \textsc{Best}@1, we use greedy decoding for the first generation and then use temperature=0.6 for the rest of the generations. We additionally report standard deviation in Table \ref{tab:main_results}.

\textbf{SFT baseline.} We create a strong SFT baseline (denoted as \methodnamesft) to compare. To select trajectories to train on, we filter by using the environment unit tests (filtering for trajectories that pass unit tests, aka rejection sampling \citep{Zelikman2022STaRBR}) We rollout using the same set of 817 tasks. We train the model for 5 epochs with a learning rate of 1e-5 and cosine learning schedule. We optimize learning rate and number of epochs via grid search. Note that we do not include any trajectories with guidance in the SFT baseline, as when comparing an SFT baseline trained with and without guidance, we find that performance is better without guidance likely due to overfitting to the guidance prompt during SFT. 

\textbf{Reward model.} To further explore test-time scaling, we train a reward model using the data generated by \methodname (including guidance). The reward model takes as input the generated patch and problem statement and produces a reward score.For the base model, we use Qwen-2.5-Coder-32B. The reward model is trained with 1e-5 learning rate for 500 steps and a warmup ratio of 0.05 with a pairwise loss. At inference time, we generate $k=32$ patches and rank them with the reward model to select one final patch for evaluation.
\nop{We train on the same amount of data used as in the RLVR experiments and filter for instances which the trajectory step is a repair step.}
\nop{At inference time, we run the reward model on the patch along with the associated problem statement to produce a reward score for each repair instance. We use $k=32$ patches during inference time generated by our scaffold.}

\textbf{Compute.} We trained on Nvidia H100 GPUs. For the 14B and 32B experiments, we use two nodes with 8 H100 GPUs each. For the 72B experiments, we use 4 nodes with 8 H100 GPUs each. The train time is 4, 6, and 10 hours for each of the 14B, 32B and 72B experiments respectively.


\subsection{Evaluation Results}

\input{tables/main-results}

\textbf{\methodname~enables models to specialize as SWE agents.} After training via \methodname, the base model (Qwen-2.5-72B-Instruct) improves from 9.4\% $\rightarrow$ 22.4\% on SWE-Bench Verified. This highlights that the agent can learn the task almost entirely through \methodname, and that the method enables weaker models to become strong agent models. This also demonstrates that environment feedback is an effective reward signal for agent training. We hypothesize that this effect would be magnified by an online RL algorithm, as online and on-policy RL generally outperforms offline DPO. We test different model sizes for \methodname, and find that performance generally scales linearly with model size.

\textbf{\methodname~vs~\methodnamesft}. We compare the RLVR trained model to an SFT baseline trained using rejection sampling. We find that the RLVR model performs better, with a score of 22.4\% for the RLVR trained model versus 20.8\% for the SFT trained model. However, it is important to note that these two are not mutually exclusive. For example, in a post-training pipeline the same trajectories can be included in the SFT pipeline, and RL data produced by \methodname~can be included during RL training for additional boosts in performance, which can be looked at as a future experiment.

\input{figures/ablations}

\textbf{Guidance is critical for \methodname}. We find that guidance is needed for successful RLVR training. Figure \ref{fig:graph_alt} highlights the gap between the model trained with and without guidance. For the non-guidance model, we perform the same training cycle but exclude any guidance-augmented data during the DPO training. We find that as \passatk~increases, the gap between the guidance and non-guidance model increases. We hypothesize that the reward landscape sparsity is a key reason for the gap between guidance and non-guidance RLVR trained models.

\textbf{RLVR data can be used as an effective test-time reward model.} Another benefit to gathering RLVR data is that it can be used to train an effective reward model. For this step, we train a reward model by using the same dataset of RLVR problems. We train on the repair patches only and include the problem statements during train and inference time. Finally, at test-time we generate 32 patches and use the reward model scores for filtering. We find that the reward model is able to discern between correct and incorrect patches at test-time and improves \passatone~from 22.4\% to 27.8\%.

\input{tables/models-compare}

%% file: tables/main-results.tex
\begin{table}[t]
\centering
\begin{tabular}{lc}
\toprule
\textbf{Model} & \textbf{\passatone (STD)} \\
\midrule
\multicolumn{2}{c}{\textbf{Base Models}} \\
\midrule
Qwen-2.5-Coder-14B & 6.8\% ($\pm$2.2) \\
Qwen-2.5-Coder-32B & 8.8\% ($\pm$2.5) \\
Qwen-2.5-72B-Instruct & 9.4\% ($\pm$2.5) \\
\midrule
\multicolumn{2}{c}{\textbf{Trained Models}} \\
\midrule
\methodname-Qwen-2.5-Coder-14B & 18.0\% ($\pm$3.4) \\
\methodname-Qwen-2.5-Coder-32B & 21.6\% ($\pm$3.5) \\
\methodnamesft-Qwen-2.5-72B-Instruct & 20.8\% ($\pm$3.5) \\
\methodname-Qwen-2.5-72B-Instruct & 22.4\% ($\pm$3.6) \\
\methodname-RM-Qwen-2.5-72B-Instruct & 27.8\% ($\pm$3.8) \\
\bottomrule
\end{tabular}
\vspace{0.4em}
\caption{Comparison between base and trained models on SWE-Bench Verified. Models trained via \methodname~are better as SWE agents, e.g. from 9.4\% $\rightarrow$ 22.4\% for Qwen-2.5-72B-Instruct. We also incorporate an SFT baseline with rejection sampling, and an additional reward model (RM) trained with RLVR data. The reward model provides additional test-time scaling (from 22.4\% $\rightarrow$ 27.8\%) while incurring minimal overhead.}
\label{tab:main_results}
\end{table}

%% file: figures/ablations.tex
\begin{figure}
    \centering
    \begin{minipage}{0.48\textwidth}
        \centering
        \includegraphics[width=\textwidth]{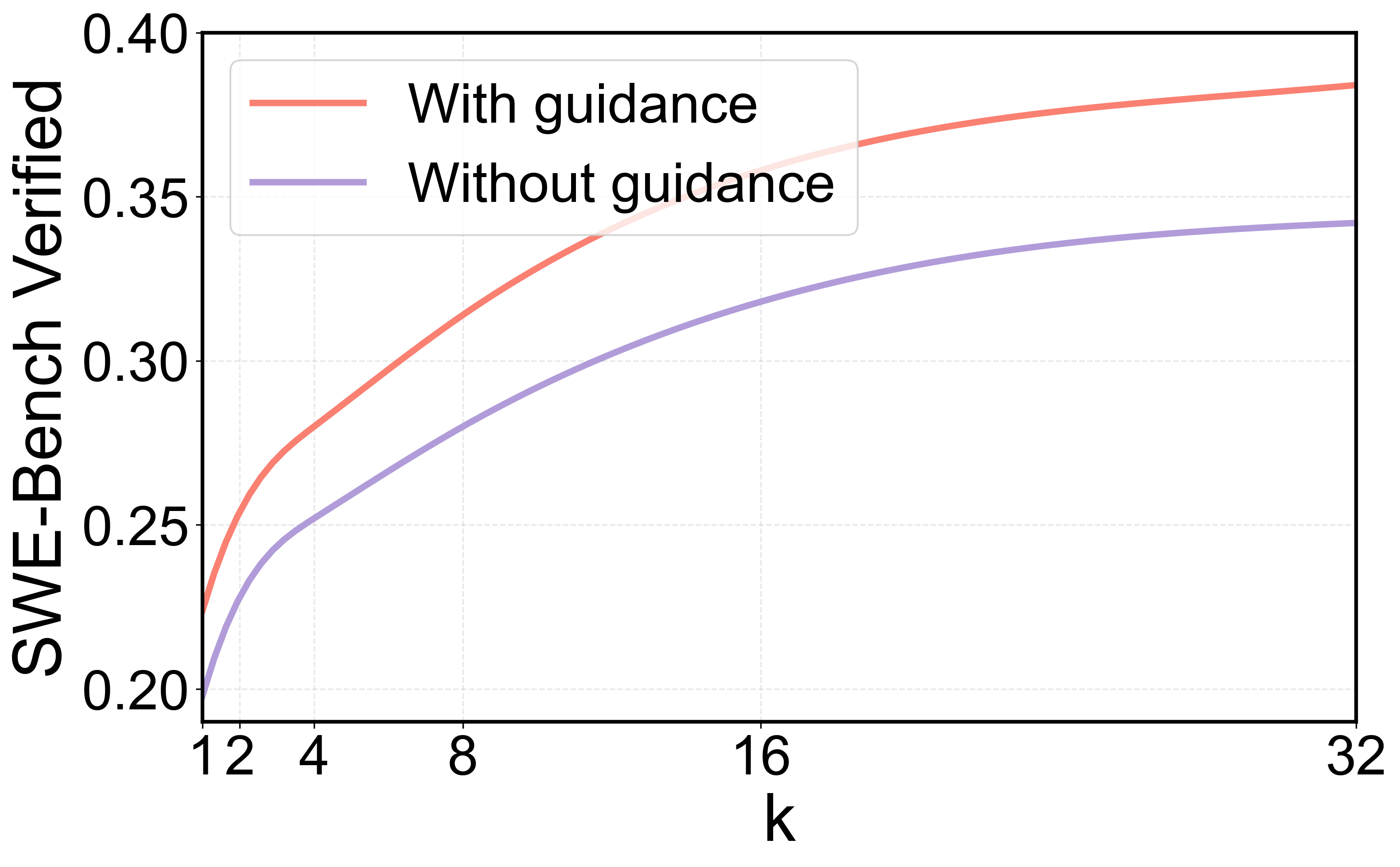}
        \caption{We compare \passatk~($k = 2^{[0, 5]}$) between the guidance and non-guidance versions of \methodname. We find that the guidance-augmented method performance is higher at \passatone~and that the performance gap increases at higher \passatk, suggesting that the guidance-trained model is able to sample more diverse and accurate generations.}
        \label{fig:graph_alt}
    \end{minipage}
    \hfill
    \begin{minipage}{0.48\textwidth}
        \centering
        \begin{tabular}{lc}
            \toprule
            \rowcolor{white}
            \textbf{Metric} & \textbf{Value} \\
            \midrule
            \multicolumn{2}{c}{\textbf{Without Guidance}} \\
            \midrule
            \rowcolor{white}
             \# of Successful Rollouts (avg) & 138 (16.9\%) \\
            Empty Patch \% & 8.71\% \\
            \midrule
            \multicolumn{2}{c}{\textbf{With Guidance}} \\
            \midrule
            \rowcolor{white}
             \# of Successful Rollouts (avg) & 165 (20.3\%) \\
            Empty Patch \% & 7.20\% \\
            \bottomrule
        \end{tabular}
        \captionof{table}{We compare trajectory rollout success rate with and without guidance. When including guidance, an average of 27 trajectories are more successful and the empty patch percent reduces by 1.51\%, highlighting that the agent is able to complete additional problems when given adequate guidance.}
        \label{tab:guidance_results}
    \end{minipage}
\end{figure}

%% file: tables/models-compare.tex
\begin{table}[t]
\centering
\begin{tabular}{lc}
\toprule
\textbf{Base Model} & \textbf{\passatone} \\
\midrule
\multicolumn{2}{c}{\textbf{Other Models}} \\
\midrule
Qwen-2.5-72B-Instruct & 9.4\% \\
SWE-Gym-Qwen-2.5-Coder-32B  \citep{Pan2024TrainingSE} (OpenHands \citep{Wang2024OpenHandsAO}) & 20.6\% \\
GPT-4o (Agentless \citep{Xia2024AgentlessDL}) & 38.8\% \\
SWE-Fixer-72B \citep{Xie2025SWEFixerTO} & 30.2\% \\
Llama3-SWE-RL-70B \citep{Wei2025SWERLAL} (Agentless \citep{Xia2024AgentlessDL}) & 41.0\% \\
DeepSeek-V3 (Agentless \citep{Xia2024AgentlessDL}) & 42.0\% \\
DeepSeek-R1 (Agentless \citep{Xia2024AgentlessDL}) & 49.2\% \\
Claude-3.5-Sonnet (Agentless \citep{Xia2024AgentlessDL}) & 50.8\% \\
\midrule
\multicolumn{2}{c}{\textbf{Our Models}} \\
\midrule
\methodname-Qwen-2.5-Coder-32B & 21.6\% \\
\methodname-Qwen-2.5-72B-Instruct & 22.4\% \\
\methodname-RM-Qwen-2.5-72B-Instruct & 27.8\% \\
\bottomrule
\end{tabular}
\vspace{0.3em}
\caption{Comparison of different models on SWE-bench Verified. \methodname~is effective while relying on a smaller training set, simpler scaffold, and/or less specialized training procedure. For example, SWE-RL \citep{Wei2025SWERLAL} has 11M training instances and SWE-Gym \citep{Pan2024TrainingSE} has 2k.}
\label{tab:models_compare}
\end{table}

%% file: sections/4-impacts.tex
\section{Broader Impacts and Limitations}

\subsection{Broader Impacts} \label{sec:impacts}

Highly capable SWE agents could have a significant impact on the tech industry. We hope that in the long term, SWE agents will serve as a collaborative tool for experienced human SWEs, eliminating tedious work and accelerating a team's ability to transform ideas into production-ready code. We believe that the demand for high-quality SWE work still outstrips the supply of SWE talent, and that advancements in training SWE agents will yield a large net social good.

\subsection{Limitations} \label{sec:limitations}

\textbf{Programming languages.} Our work is limited to Python as it is the most widely used programming language, and is supported by other infrastructure components such as SWE-Bench and Agentless. We leave other languages for future work.

\textbf{Dataset limitations.} Our training data is limited to a relatively small set of repositories due to cost considerations. Additionally, there is (to our knowledge) only one substantial test set for SWE agents that has been human-validated for reliability. It's possible that the dataset contains contamination with respect to pre-training \citep{Zhang2024ACE}, although we ensure that repositories do not overlap between training and test sets.

\textbf{Online versus offline RL.} In this work, we explore \methodname as an offline algorithm. However, it is possible to apply the same methods to online RL. For example, \methodname could run for several iterations with GRPO or PPO. Previous work \citep{Lambert2024TLU3P, DeepSeekAI2025DeepSeekR1IR, DeepSeekAI2024DeepSeekV3TR} has found that online RL is an effective way to train language models for reasoning and coding tasks. Since this requires additional infrastructure setup for the online environment, we focus on offline RL for this work.

%% file: sections/5-relatedwork.tex
\section{Related Work}

\subsection{Training software engineering agents}

Our work mainly focuses on training software engineering agents, for instance, agents able to autonomously solve issues and debug coding problems in a large repository, which was first defined by SWE-Bench \citep{Jimenez2023SWEbenchCL}. These agents are assessed on their ability to produce a patch that passes a set of unit tests, and problem statements are derived from previous pull-requests or commits in that repo. More recently, commit0 \citep{Zhao2024Commit0LG} assesses agent ability to create an repository from scratch.

Early software engineering agent designs involved improvements on the agent scaffold, with a fixed model and no additional training. SWE-Agent \citep{Yang2024SWEagentAI} uses an agent-computer interface to allow the agent to iteratact with the environment via tool-use. Agentless \citep{Xia2024AgentlessDL} takes a multi-step approach with localization, repair, and validation steps. React \citep{Yao2022ReActSR} incorporates tool-use with reasoning traces to generate agent trajectories.

Several works have explored training software engineering agents by keeping the scaffold constant and training the model itself. SWE-Gym \citep{Pan2024TrainingSE} trained the model via rejection sampling, and also explores test-time scaling for model patches. SWE-RL \citep{Wei2025SWERLAL} uses reinforcement learning on Github repos via GRPO and a cosine similarity scoring function. SWE-Fixer \citep{Xie2025SWEFixerTO} trained a code retrieval and code editing module for the localization and repair steps respectively.

\subsection{Reinforcement Learning from Verifiable Rewards}

RLVR (coined by \citep{Lambert2024TLU3P}) denotes a set of research where RLHF training methods (such as PPO \citep{Schulman2017ProximalPO}, DPO \citep{Rafailov2023DirectPO}, and GRPO \citep{Shao2024DeepSeekMathPT}) are applied using verifiable feedback as a reward signal. Typical RLHF pipelines involve collecting a set of human preferences to train a reward model, however, in RLVR the reward model is not used and instead the reward score is via a grader or set of unit tests. DeepSeek r1 \citep{DeepSeekAI2025DeepSeekR1IR} shows the utility of scaling test-time thoughts, which are generated by the model during the reinforcement learning process. In other post-training pipelines, RLVR is a critical part of the training process. For example, in Tulu3, the last stage of the post-training pipeline is an RLVR stage \citep{Lambert2024TLU3P}. In DeepSeek v3, the model goes through an RLVR stage for code and reasoning training \citep{DeepSeekAI2024DeepSeekV3TR}. In terms of guidance, PlanSearch \citep{Wang2024PlanningIN} shows that including a plan helps to improve search diversity in coding tasks.

\subsection{Progress in large language model agents}

Apart from coding agents, there has been a significant amount of progress in LM agents. Similar to SWE-Bench, several benchmarks exist for enviroment-based computer-use and web agents. Some examples are OSWorld \citep{Xie2024OSWorldBM} which provides environment infrastructure for computer-use agents, Mind2Web \citep{Deng2023Mind2WebTA} which evaluates agents on a set of open-ended website tasks, and ToolComp \citep{Nath2025ToolCompAM} for tool-use. Works that involve training models for agentic tasks are more sparse, with UI-Tars \citep{Qin2025UITARSPA} training a model using screenshots as input for GUI agents. More recent works, such as OpenAI Operator and WebRL \citep{Qi2024WebRLTL} show the effectiveness of training a model for computer-use and web-use respectively.

%% file: sections/6-conclusion.tex
\section{Conclusion}

In this paper, we discuss \methodname, a pipeline for training language models via RLVR. We introduce agent guidance, which enables efficient use of provided environments and reduces sparsity in reward signal landscapes for agents via agent self-improvement from additional environment exploration. We prove our method on SWE-Agents, achieving improved performance by training via RL and beating SFT baselines. Through analysis of test-time scaling, we find that the guidance gap widens at higher \passatk~and that the guidance-based RL data can be used to train a reward model for test-time patch selection. Our findings indicate the promise of RL for agent training and reveal potential for future avenues of environment-based RL training.

%% file: sections/appendix.tex
\section{Technical Appendices and Supplementary Material}

\begin{table}[H]
\centering
\begin{tabular}{lc}
\toprule
\textbf{Method} & \textbf{Score} \\
\midrule
SFT with guidance & 16.8 \\
SFT without guidance & 20.8 \\
\bottomrule
\end{tabular}
\vspace{0.3em}
\caption{Comparison of SFT with and without guidance.}
\label{tab:sft_comparison}
\end{table}

\begin{figure}[htbp]
    \centering
    \begin{framed}
    \lstset{
        basicstyle=\ttfamily\small,
        breaklines=true,
        columns=flexible,
        keepspaces=true,
        showstringspaces=false,
        frame=none
    }
    \begin{lstlisting}
You are an expert software engineer helping to generate helpful hints for debugging and fixing issues.
REPOSITORY: {repo}
PROBLEM STATEMENT:
{problem_statement}
PATCH:
{patch}
STACKTRACE:
{stacktrace_hint}
Based on the information above and the following hints, create a concise, helpful hint that would guide a developer to fix the issue efficiently:
The hint should be in the following format:
PLAN HINT: Explain what steps need to be taken based on the problem statement. Be specific and actionable, focusing on the core issue. Be concise yet comprehensive (200-300 words maximum)
ENVIRONMENT FEEDBACK HINT: Explain what needs to be fixed based on the stacktrace. Be specific and actionable, focusing on the core issue. Be concise yet comprehensive (200-300 words maximum)
ENVIRONMENT INTERACTION HINT: Clearly identify potential file(s) and location(s) need to be modified.
Format your hint with clear section headings (PLAN HINT, ENVIRONMENT FEEDBACK HINT, ENVIRONMENT INTERACTION HINT).
Each part should have 1-2 sentences, except for ENVIRONMENT INTERACTION which should just be a location-based hint (file paths, and usually line number, function name, etc.).
    \end{lstlisting}
    \end{framed}
    \caption{Guidance generation prompt. \texttt{repo} refers to the name of the repo, \texttt{problem\_statement} is the task statement, \texttt{patch} is the reference patch, and \texttt{stacktrace\_hint} is the stacktrace from the previous attempt.}

    \label{fig:guidance-template}
\end{figure}

\begin{figure}[htbp]
    \centering
    \begin{framed}
    \lstset{
        basicstyle=\ttfamily\small,
        breaklines=true,
        columns=flexible,
        keepspaces=true,
        showstringspaces=false,
        frame=none
    }
    \begin{lstlisting}
    In addition, please carefully consider the following hint, which includes a suggested path(s) to file(s) that may be problematic.
    ### Hint ###
    {hint}
    ###
    Please solve the problem starting from the very beginning. Use the hint as guidance, but do not assume any steps have already been completed. When using the hint, you still need to explain your thought process as if you did not have access to the hint.
    \end{lstlisting}
    \end{framed}
    \caption{Example of including guidance during trajectory generation, in which the above prompt is appended to the trajectory.}
    \label{fig:additional-guidance}
\end{figure}